\documentclass{article}
\usepackage[utf8]{inputenc}

\usepackage[dblblindworkshop, final]{neurips_2025}

\usepackage[T1]{fontenc}
\usepackage{amsmath, amssymb, amsthm}
\usepackage{mathtools}
\usepackage{bbm}
\usepackage{dsfont}
\usepackage[verbose=true,letterpaper]{geometry}
  \newgeometry{
    textheight=9in,
    textwidth=5.5in,
    top=1in,
    headheight=12pt,
    headsep=25pt,
    footskip=30pt
  }

\usepackage[dvipsnames]{xcolor}
\definecolor{shadecolor}{gray}{0.9}

\usepackage{graphicx}
\usepackage[labelfont=bf]{caption}
\usepackage[format=hang]{subcaption}

\usepackage{booktabs, array}
\usepackage{multirow}

\usepackage{algorithm}
\usepackage{algpseudocode}
\algrenewcommand\algorithmicrequire{\textbf{Input:}}
\algrenewcommand\algorithmicensure{\textbf{Output:}}
\usepackage{listings}
\usepackage{fancyvrb}
\fvset{fontsize=\small}

\usepackage{natbib}
\usepackage[colorlinks,linktoc=all]{hyperref}
\usepackage[all]{hypcap}
\hypersetup{citecolor=MidnightBlue}
\hypersetup{linkcolor=MidnightBlue}
\hypersetup{urlcolor=MidnightBlue}
\usepackage[nameinlink,capitalise]{cleveref}
\creflabelformat{equation}{#2\textup{#1}#3}  

\setcitestyle{authoryear,round,citesep={;},aysep={,},yysep={;}}

\lstdefinestyle{mystyle}{
    commentstyle=\color{OliveGreen},
    numberstyle=\tiny\color{black!60},
    stringstyle=\color{BrickRed},
    basicstyle=\ttfamily\scriptsize,
    breakatwhitespace=false,
    breaklines=true,
    captionpos=b,
    keepspaces=true,
    numbers=none,
    numbersep=5pt,
    showspaces=false,
    showstringspaces=false,
    showtabs=false,
    tabsize=2
}
\usepackage{siunitx}
\usepackage{textcomp}
\usepackage{wrapfig}

\newsavebox\CBox 

\def\UL#1{\underline{#1}}

\theoremstyle{plain}

\workshoptitle{Frontiers in Probabilistic Inference: Sampling Meets Learning}
\title{Improving Constrained Language Generation via Self-Distilled Twisted Sequential Monte Carlo}

\author{
  Sooyeon Kim \\
  Seoul National University \\
  \texttt{ksooyeon3@snu.ac.kr} \\
  \And
  Giung Nam, Byoungwoo Park, Juho Lee \\
  Korea Advanced Institute of Science and Technology \\
  \texttt{\{giung,\;bw.park,\;juholee\}@kaist.ac.kr} \\
}

\begin{document}
\maketitle

\begin{abstract}
\noindent
Recent work has framed constrained text generation with autoregressive language models as a probabilistic inference problem.
Among these, \citet{zhao2024probabilistic} introduced a promising approach based on twisted Sequential Monte Carlo, which incorporates learned twist functions and twist-induced proposals to guide the generation process.
However, in constrained generation settings where the target distribution concentrates on outputs that are unlikely under the base model, learning becomes challenging due to sparse and uninformative reward signals.
We show that iteratively refining the base model through self-distillation alleviates this issue by making the model progressively more aligned with the target, leading to substantial gains in generation quality.
\end{abstract}

\section{Introduction}

Recent progress in large language models has been primarily driven by autoregressive modeling; these models are typically optimized to predict a next-token given its preceding context~\citep{brown2020language}.
Specifically, modern autoregressive language models estimate the conditional probability of a sequence $s_{1:T}$ given an initial prompt $s_{0}$ as
\begin{align}\label{eq:p_lm}\textstyle
p_{\text{LM}}(s_{1:T} | s_{0}) = \prod_{t=1}^{T} p_{\text{LM}}(s_t | s_{0:t-1}),
\end{align}
where $s_{t}$ is the $t^{\text{th}}$ token and $s_{0:t-1}$ denotes the sequence of tokens up to step $t-1$, including the initial prompt $s_{0}$.
Assuming access to a potential function $\phi$, which scores the preference over full sequences as $\phi : s_{1:T} \mapsto \mathbb{R}^{+}$, the target distribution $\sigma$ from which we aim to sample in \textit{constrained language generation} setups (along with its unnormalized form $\tilde{\sigma}$) can be defined as
\begin{align}\label{eq:sigma}\textstyle
\sigma(s_{1:T} | s_{0}) = {\tilde{\sigma}(s_{1:T}|s_{0})} / {\mathcal{Z}_{\sigma}(s_{0})},
\text{ where }
\tilde{\sigma}(s_{1:T} | s_{0}) = p_{\text{LM}}(s_{1:T} | s_{0}) \phi(s_{1:T}),
\end{align}
and $\mathcal{Z}_{\sigma}(s_{0}) = \sum_{s_{1:T}} p_{\text{LM}}(s_{1:T}|s_{0}) \phi(s_{1:T})$ denotes the normalization constant, which is intractable to compute in practice due to the summation over all possible sequences.
Given that the target $\tilde{\sigma}$ is unnormalized but tractable to evaluate for individual sequences, constrained language generation can be naturally framed as a \emph{probabilistic inference} problem~\citep{korbak2022rl,lew2023sequential,zhao2024probabilistic,loula2025syntactic}, where the goal is to sample 
$s_{1:T} \sim \sigma(\cdot | s_{0})$.

A primary challenge in sampling from $\sigma$ lies in its \emph{non-causal} structure.
While $p_{\text{LM}}$ permits efficient left-to-right sampling, $\sigma$ depends on the full sequence score $\phi(s_{1:T})$.
Consequently, computing the marginal $\sigma(s_{1:t} | s_{0})$ requires summing over all possible future continuations:
\begin{align}\textstyle
\sigma(s_{1:t}|s_{0}) \propto \sum_{s_{t+1:T}} p_{\text{LM}}(s_{t+1:T} | s_{0:t}) \phi(s_{1:T}),
\end{align}
which is intractable in practice due to the exponential number of possible continuations.
To address this, \citet{zhao2024probabilistic} proposed leveraging the \emph{Twisted Sequential Monte Carlo (TSMC)} framework~\citep{heng2020controlled}, which generalizes importance sampling by introducing a sequence of intermediate distributions $\{\pi_{t}(s_{1:t}|s_{0})\}_{t=1}^{T-1}$ that progressively approach the final target distribution $\pi_{T}(s_{1:T}|s_{0}) = \sigma(s_{1:T}|s_{0})$.
In the TSMC framework, each $\pi_{t}$ (along with its unnormalized form $\tilde{\pi}$) is represented as a \emph{twisted intermediate target distribution}, defined by applying a \emph{twist} function $\psi_{t} : s_{1:t} \mapsto \mathbb{R}$ to the base language model $p_{\text{LM}}$:
\begin{align}\textstyle
\pi_{t}(s_{1:t} | s_{0}) = {\tilde{\pi}_{t}(s_{1:t}|s_{0})} / {\mathcal{Z}_{\pi_{t}}(s_{0})},
\text{ where }
\tilde{\pi}_{t}(s_{1:t}|s_{0}) = p_{\text{LM}}(s_{1:t} | s_{0}) \psi_{t}(s_{1:t}),
\end{align}
and $\mathcal{Z}_{\pi_{t}}(s_{0})$ is the normalization constant.
If we approximate $\psi_{t}(s_{1:t})$ using a shared neural network $\psi_{\theta}(s_{1:t})$ parameterized by $\theta$, we have the approximate intermediate target distribution of $\pi_{t,\theta}(s_{1:t}|s_{0}) \propto p_{\text{LM}}(s_{1:t}|s_{0}) \psi_{\theta}(s_{1:t})$.
The Contrastive Twist Learning~\citep[CTL;][]{zhao2024probabilistic} framework trains this neural twist function by minimizing the loss:
\begin{align}\textstyle\label{eq:ctl_loss}
\mathcal{L}_{\text{CTL}}(\theta)
= \sum_{t=1}^{T} D_{\text{KL}}(\sigma(s_{1:t}|s_{0}) || \pi_{t,\theta}(s_{1:t}|s_{0})),
\end{align}
which yields the following negative gradient with respect to the parameters $\theta$:
\begin{align}\textstyle\label{eq:ctl_gradient}
-\nabla_{\theta}\mathcal{L}(\theta)
= \sum_{t=1}^{T} \left\{
    \mathbb{E}_{\sigma(s_{1:t}|s_{0})} \left[ \nabla_{\theta} \log{\psi_{\theta}(s_{1:t})} \right]
    - \mathbb{E}_{\pi_{t,\theta}(s_{1:t}|s_{0})} \left[ \nabla_{\theta} \log{\psi_{\theta}(s_{1:t})} \right]
\right\}.
\end{align}
Using $\psi_{\theta}$ and the corresponding \emph{twist-induced proposal} $q_{t}(s_{t} | s_{0:t-1}) \propto p_{\text{LM}}(s_{t} | s_{0:t-1}) \psi_{\theta}(s_{1:t})$,
which aims to minimize the variance of the importance weights~\citep{zhao2024probabilistic},
we perform the following three steps iteratively for each time step $t=1,\dots,T$, starting from $K$ particles $\{s_{0}^{k}\}_{k=1}^{K}$:
\begin{enumerate}
\item \label{enu:extending} \emph{(Extending)} Sample $s_{t}^{k} \sim q(\cdot | s_{0:t-1}^{k})$ and extend $s_{0:t}^{k} \gets \texttt{concat}(s_{0:t-1}^{k}, s_{t}^{k})$.
\item \label{enu:reweighting} \emph{(Reweighting)} Compute
$w_{t}^{k} \gets \sum_{s_{t}} p_{\text{LM}}(s_{t}|s_{0:t-1}) \psi_{\theta}(s_{1:t}) / \psi_{\theta}(s_{1:t-1})$ for $t<T$, and at the final time step $t=T$, $w_{t}^{k} \gets \sum_{s_{T}} p_{\text{LM}}(s_{T}|s_{0:T-1}) \phi(s_{1:T}) / \psi_{\theta}(s_{1:T-1})$.
\item \emph{(Resampling)} Sample $\{k_{i}\}_{i=1}^{K} \overset{\text{i.i.d.}}{\sim} \text{Categorical}(\{\bar{w}_{t}^{i}\}_{i=1}^{K})$, where $\bar{w}_{t}^{i} =  w_{t}^{i} / \sum_{j=1}^{K}w_{t}^{j}$, and reassign $(s_{0:t}^{i})_{i=1}^{K} \gets (s_{0:t}^{k_{i}})_{i=1}^{K}$.
\end{enumerate}
Throughout the paper, we denote the TSMC sampling procedure as $s_{1:T} \sim \mathrm{TSMC}(p_{\text{LM}}, \psi_{\theta})$, where the number of particles $K$ is treated as a hyperparameter.
Although TSMC is theoretically well-founded, being rooted in the Feynman–Kac formalism~\citep{del2004feynman} and supported by various guarantees~\citep{doucet2001sequential,chopin2020introduction,heng2020controlled}, its practical effectiveness can be limited. A poorly chosen twist often leads to rapid particle degeneracy, and the repeated evaluation of $\psi_{\theta}(s_{1:t})$ during sampling introduces a computational bottleneck. As a result, large-capacity neural networks cannot be used for $\psi_{\theta}$ despite their potential to yield more effective twists, creating a fundamental trade-off between expressiveness and computational feasibility.
Indeed, twist learning via CTL becomes particularly challenging in constrained text generation tasks, where the target distribution emphasizes outputs that are unlikely under the base model, especially when $\psi_{\theta}$ is implemented as a lightweight multilayer perceptron (MLP) for runtime efficiency (cf. \cref{sec:exp}).

\section{Self-Distilled Twisted Sequential Monte Carlo}

\cref{alg:short} outlines our proposed \emph{self-distilled TSMC} procedure.
The process starts from the initial pair $\smash{(p_{\text{LM}}^{(0)}, \psi_{\theta}^{(0)})}$, obtained using the original CTL framework of \citet{zhao2024probabilistic} (line 1), which we regard as the 0-th generation.
The base model is then iteratively refined through self-distillation (line 3), and the corresponding twist function is updated via a modified CTL approach (line 4), producing a sequence of progressively improved pairs $\smash{(p_{\text{LM}}^{(m)}, \psi_{\theta}^{(m)})}$ across generations.
In what follows, we detail the steps involved in each phase and describe how the CTL framework is adapted to incorporate a refined base model obtained through self-distillation.

\begin{algorithm}[t]
\caption{Self-distilled TSMC}\label{alg:short}
\begin{algorithmic}[1]
    \Require{Base model $p_{\text{LM}}^{(0)}$, potential function $\phi$, and the number of generations $M$.}
    \Ensure{Refined base model $p_{\text{LM}}^{(M)}$ and corresponding twist function $\psi_{\theta}^{(M)}$.}
    \State{$\psi_{\theta}^{(0)} \xleftarrow[]{\text{CTL}} p_{\text{LM}}^{(0)}$.}
    \For{$m=1,\dots,M$}
        \State{$p_{\text{LM}}^{(m)} \xleftarrow[]{\text{SD}} \operatorname{TSMC}(p_{\text{LM}}^{(m-1)}, \psi_{\theta}^{(m-1)})$.}
        \State{$\psi_{\theta}^{(m)} \xleftarrow[]{\text{CTL-$m$}} p_{\text{LM}}^{(m)}$.}
    \EndFor
\end{algorithmic}
\end{algorithm}

\textbf{Phase \#1: Self-distillation.}
Distillation in generative language models is fairly straightforward: the student model is optimized to maximize the likelihood of sequences output by the teacher model~\citep{kim2016sequence}.
In our self-distillation phase, the base model of the $m$-th generation is trained on TSMC samples from the $(m-1)$-th generation:
\begin{align}\textstyle
p_{\text{LM}}^{(m)}(\cdot|s_{0}) = \underset{p_{\text{LM}}(\cdot|s_{0})}{\arg\min} \; \mathbb{E}_{
    s_{1:T} \sim \mathcal{S}
} \left[
    -\log{p_{\text{LM}}(s_{1:T}|s_{0})}
\right].
\end{align}
where $\mathcal{S}$ represents the set of text samples generated by $\mathrm{TSMC}(p_{\text{LM}}^{(m-1)}, \psi_{\theta}^{(m-1)})$.

\textbf{Phase \#2: (Modified) contrastive twist learning.}
A key aspect of our self-distilled TSMC procedure is that the base model progressively evolves, moving away from the initial $p_{\text{LM}}$ and gradually approaching the target distribution $\sigma$ over successive generations.
Intuitively, this facilitates smoother training of the twist function (which we also demonstrate empirically in the following section), yet it simultaneously introduces a complication that requires modification in implementing CTL.
Specifically, from the perspective of the base model in the $m$-th generation, the target distribution $\sigma$ can be rewritten as:
\begin{align}\textstyle
\sigma(s_{1:T}|s_{0})
\propto p_{\text{LM}}^{(m)}(s_{1:T}|s_{0}) \phi^{(m)}(s_{1:T}),
\end{align}
where
$\phi^{(m)}(s_{1:T}) \coloneqq p_{\text{LM}}^{(0)}(s_{1:T}|s_{0}) \phi(s_{1:T}) / p_{\text{LM}}^{(m)}(s_{1:T})$.
Here, $\phi^{(m)}$ acts as an effective potential in the $m$-th generation, adjusting the current base model's density to align with $\sigma$ defined under $p_{\text{LM}}$.
As a result, the approximate positive sampling procedure used to compute the first term in \cref{eq:ctl_gradient} must be adjusted accordingly, since the importance weights are now governed by the generation-specific effective potential $\phi^{(m)}(s_{1:T})$ over full sequences.
The modified procedure becomes:
\begin{enumerate}
    \item Using SMC with proposal $q$, generate candidates
    $$\textstyle
    \{s_{1:T}^{k}\}_{k=1}^{K} \overset{\text{i.i.d.}}{\sim} q(s_{1:T} | s_{0}).
    $$
    \item For each $k=1,\dots,K$, compute modified importance weights
    $$\textstyle
    w^{(m)}(s_{1:T}^{k}) = p_{\text{LM}}^{(m)}(s_{1:T}^{k}|s_{0}) \phi^{(m)}(s_{1:T}^{k}) / q(s_{1:T}^{k}|s_{0}).
    $$
    \item Using $\hat{w}^{(m)}(s_{1:T}^{k}) = w^{(m)}(s_{1:T}^{k}) / \sum_{j=1}^{K} w^{(m)}(s_{1:T}^{j})$ for $k=1,\dots,K$, approximate
    \begin{align*}\textstyle
    \mathbb{E}_{\sigma(s_{1:t}|s_{0})} \left[ \nabla_{\theta}\log{\psi_{\theta}^{(m)}(s_{0:t})} \right]
    \approx \sum_{k=1}^{K} \hat{w}^{(m)}(s_{1:T}^{k}) \nabla_{\theta} \log{\psi_{\theta}^{(m)}(s_{1:t}^{k}))}.
    \end{align*}
\end{enumerate}
Notably, this complication does not affect the Extending step (\cref{enu:extending}) of the TSMC sampling.
Even in the original CTL setup where $p_{\text{LM}}$ serves as the base model, the twist-induced proposal remains intractable to evaluate exactly at the final timestep, as computing the terminal potential for all possible $s_{1:T}$ given $s_{1:T-1}$ is prohibitively costly.
As such, we continue to rely on the approximation $\psi_{\theta}^{(m)}(s_{1:T}) \approx \phi^{(m)}(s_{1:T})$ to enable practical proposal sampling.

\section{Experiments}
\label{sec:exp}

\textbf{Setup.}
We focus on toxic story generation; the base language model $p_{\text{LM}}(\cdot|s_{0})$ generates continuations $s_{1:T}$ from the initial prompt $s_{0} = \text{`Once upon a time, there was a'}$, and we define
$$\textstyle
\phi(s_{1:T})
\coloneqq p(\text{toxic}|s_{1:T})^{\beta}
= \exp{\left\{\beta\cdot\log{p(\text{toxic}|s_{1:T})}\right\}},
$$
where $\beta \geq 0$ controls the strength of the toxicity signal.
For both $p_{\text{LM}}$ and $\phi$, we adopted publicly available pre-trained models (see \cref{app:models}).
To evaluate whether the generated texts are both diverse and toxic, we employ two metrics:
\emph{1) Toxicity} is defined as $p(\text{toxic} \mid s_{1:T})$, following our definition of the potential function.
\emph{2) Similarity} is defined as the average pairwise cosine similarity of sentence embeddings obtained from Sentence Transformers~\citep{reimers2019sentence}.
From early experiments with rejection sampling (see \cref{app:exact}), we observed that meaningful toxic story generation occurs when $\beta = 10.0$, a regime that effectively captures the sparse reward problem.
Therefore, our main results are reported under the target distribution $\sigma$ with $\beta = 10.0$.

\begin{wrapfigure}{r}{0.5\textwidth}
    \centering
    \includegraphics[width=\linewidth]{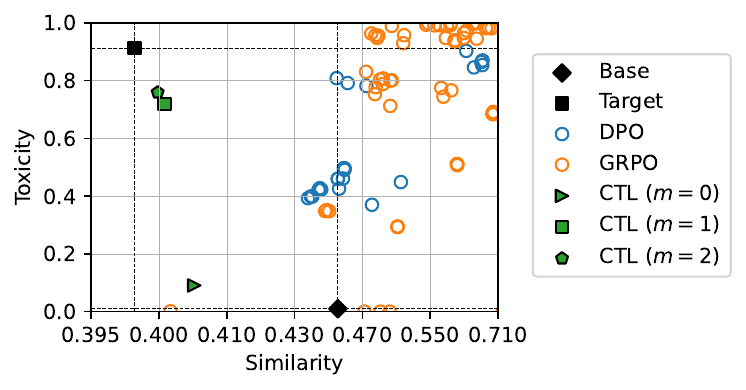}
    \vspace{-2em}
    \caption{Similarity-Toxicity plot.\vspace{0.5em}}
    \label{fig:main}
    \vspace{-1.0em}
\end{wrapfigure}

\textbf{Results.}
\cref{fig:main} illustrates a Similarity–Toxicity scatter plot.
The upper-left region represents a favorable area where the model generates samples that are both diverse and toxic.
We employ \( K = 100 \) particles for the TSMC sampling procedure during training, and \( K = 50 \) during testing.
\emph{Base} and \emph{Target} refer to the pre-trained model \( p_{\text{LM}} \) and the target distribution \( \sigma \), respectively.
Samples generated by the proposed self-distilled TSMC approach progressively approach the \emph{Target} distribution across iterations, surpassing practical baselines that directly fine-tune the pre-trained \( p_{\text{LM}} \) using \( \phi \) to approximate \( \sigma \), such as \emph{Direct Preference Optimization~\citep[DPO;][]{rafailov2023direct}} and \emph{Group Relative Policy Optimization~\citep[GRPO;][]{shao2024deepseekmath}}.

\begin{wrapfigure}{r}{0.5\textwidth}
    \vspace{-1.0em}
    \centering
    \includegraphics[width=\linewidth]{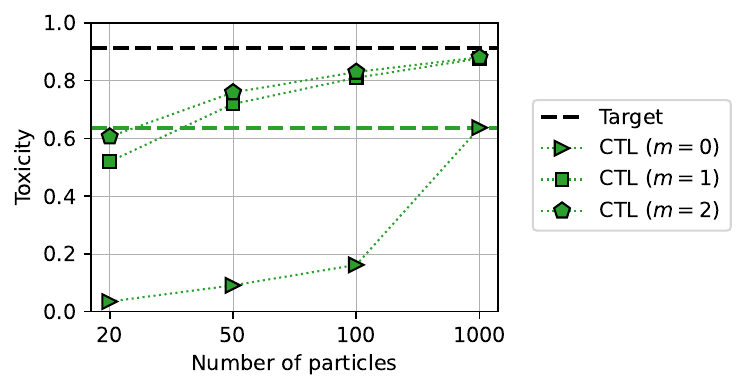}
    \vspace{-2em}
    \caption{Particle efficiency plot.\vspace{0.5em}}
    \label{fig:num_particles}
    \vspace{-1.5em}
\end{wrapfigure}

\textbf{Ablation of particle efficiency.}
To further assess the quality of the learned twist, we measure the toxicity of TSMC samples while varying the number of particles $K \in \{20, 50, 100, 1000\}$.
\cref{fig:num_particles} shows that as the generation index $m$ increases, the toxicity progressively approaches that of the \textit{Target} (indicated by the black dashed line).
Our approach yields competitive performance even with a small number of particles, using the refined base model and the corresponding learned twist function.
These results demonstrate that our approach iteratively enhances particle efficiency by progressively aligning the base model with the target, enabling high-quality generation with fewer particles.

\textbf{Evaluating twist-induced proposal.}
Following \citet{zhao2024probabilistic}, we further compute the KL divergence $D_{\text{KL}}( \sigma(s_{1:T} | s_{0}) \,\|\, q(s_{1:T} | s_{0}) )$ between the target $\sigma(s_{1:T} | s_{0})$ and the twist-induced proposal over full sequences $q(s_{1:T} | s_{0}) = \prod_{t=1}^{T} q_{t}(s_{t} | s_{0:t-1})$.
This divergence quantifies how well the proposal approximates the target, thereby serving as a measure of the quality of twist learning.
As the base model is progressively refined ($p_{\text{LM}}^{(0)} \rightarrow p_{\text{LM}}^{(1)} \rightarrow p_{\text{LM}}^{(2)}$), the divergence steadily decreases across iterations ($7.971 \rightarrow 7.030 \rightarrow 7.016$), indicating that the learned proposal becomes increasingly aligned with the target distribution.

\section{Conclusion}

In this work, we identify a key limitation of TSMC in practical constrained text generation settings, where the reward signal is often significantly misaligned with the base language model.
When such misalignment occurs, twist learning with a moderate-sized neural network often fails to yield a sufficiently representative particle system that captures the target distribution, leading to degraded sampling performance.
To address this, we propose a self-distillation-based framework that iteratively refines the base model.
This process facilitates more effective twist learning and leads to substantial improvements in TSMC sampling quality, even with a simple MLP twist and a small number of particles.
Promising future directions include compressing the iterative refinement process and designing more effective twist learning methods, thereby enhancing the scalability and practicality of TSMC for modern large language models.

\bibliographystyle{plainnat}
\bibliography{reference}

\newpage
\appendix

\section{Pre-trained models}
\label{app:models}

Below is a summary of the pre-trained weights used in our experiments, all of which are publicly available on the Hugging Face Hub~\citep{wolf2020transformers}.\footnote{\href{https://huggingface.co/}{https://huggingface.co/}}
\begin{itemize}
    \item \texttt{TinyStories-33M} ($p_{\text{LM}}$): \href{https://huggingface.co/roneneldan/TinyStories-33M}{roneneldan/TinyStories-33M}, licensed under MIT License.\footnote{\href{https://mit-license.org/}{https://mit-license.org/}}
    \item \texttt{ToxicityModel} ($\phi$): \href{https://huggingface.co/nicholasKluge/ToxicityModel}{nicholasKluge/ToxicityModel}, licensed under Apache License 2.0.\footref{apache-2.0}
    \item \texttt{all-MiniLM-L6-v2}: \href{https://huggingface.co/sentence-transformers/all-MiniLM-L6-v2}{sentence-transformers/all-MiniLM-L6-v2}, licensed under Apache License 2.0.\footnote{\href{https://www.apache.org/licenses/LICENSE-2.0}{https://www.apache.org/licenses/LICENSE-2.0}\label{apache-2.0}}
\end{itemize}

\section{Supplementaries for rejection sampling experiments}
\label{app:exact}

Given that the potential function $\phi$ satisfies $\forall s_{1:T} : \phi(s_{1:T}) \in [0, 1]$, \emph{rejection sampling} can be employed to obtain exact samples from the target distribution $\sigma$.
Specifically, candidate sequences $\hat{s}_{1:T}$ are drawn from $p_{\text{LM}}(\cdot|s_{0})$, and each candidate is accepted with probability proportional to $\phi(\hat{s}_{1:T})$.
Since the candidates are sampled from $p_{\text{LM}}$ during rejection sampling, the acceptance ratio serves as a reliable metric to reveal the sparse reward problem we aim to simulate, which manifests when the toxicity constraint is strongly enforced, i.e., for $\beta > 1$.

\begin{figure}[ht!]
  \centering
  \begin{minipage}[b]{0.48\textwidth}
    \centering
    \includegraphics[width=\linewidth]{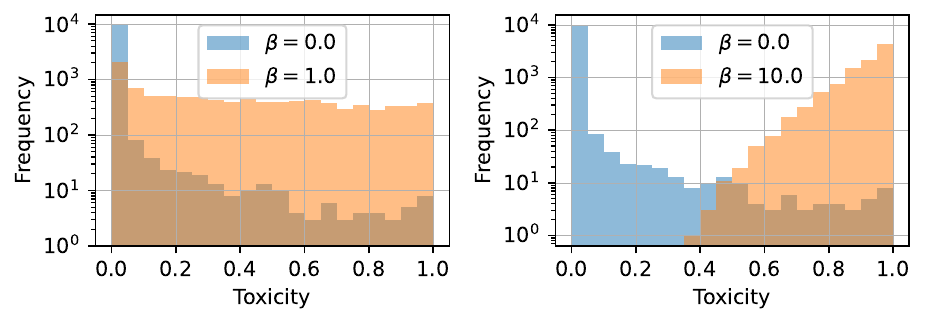}
    \caption{Toxicity histograms.}
    \label{fig:exact_main}
  \end{minipage}
  \hfill
  \begin{minipage}[b]{0.48\textwidth}
    \centering
    \resizebox{\linewidth}{!}{%
      \begin{tabular}{cccc}
        \toprule
        $\beta$ & Accept ratio ($\uparrow$) & Toxicity ($\uparrow$) & Similarity ($\downarrow$) \\
        \midrule
        $\phantom{0}0.0$ & 100.00 \%           & 0.010 & 0.456 \\
        $\phantom{0}1.0$ & \phantom{00}0.99 \% & 0.378 & 0.405 \\
        $10.0$ & \phantom{00}0.06 \%           & \UL{0.913} & \UL{0.398} \\
        \bottomrule
      \end{tabular}%
    }
    \vspace{1.5em}
    \captionof{table}{Rejection sampling results.}
    \label{tab:exact}
  \end{minipage}
\end{figure}

\cref{fig:exact_main} shows toxicity histograms of the samples obtained via rejection sampling, highlighting the substantial gap between $p_{\text{LM}}$ (i.e., $\beta=0.0$) and $\sigma$, particularly when $\beta=10.0$ is used; note that the y-axis is in log scale.
\cref{tab:exact} further demonstrates that the target distribution with $\beta=1.0$ fails to generate highly toxic samples, while $\beta=10.0$ produces significantly more toxic outputs, with mean toxicities of 0.3779 and 0.9125, respectively.
However, it comes at a price of a substantially reduced acceptance ratio during the rejection sampling procedure, implying the severity of the sparse reward problem where the target distribution concentrates on low-probability, high-toxicity regions relative to the base model.
\cref{app:exact} further presents qualitative examples obtained via rejection sampling from $\sigma$ with $\beta \in \{0.0, 1.0, 10.0\}$.

\cref{table/exact_example0.0,table/exact_example1.0,table/exact_example10.0} show generated examples obtained via rejection sampling from the target distributions with $\beta = 1.0$ and $\beta = 10.0$, respectively.
While we observe that the \texttt{ToxicityModel} often assigns high toxicity scores in response to specific keywords---sometimes overlooking the overall semantic meaning of the sentence---this is not a critical issue, as our experimental setup adopts the \texttt{ToxicityModel} as the ground truth scoring function by design.

\begin{table}[ht!]
\centering
\caption{Representative examples sampled from the target with $\beta=0.0$.}
\label{table/exact_example0.0}
\scriptsize
\setlength{\tabcolsep}{8pt}
\renewcommand{\arraystretch}{1.5}
\makebox[\textwidth][c]{
\begin{tabular}{|c|p{12cm}|}
\hline
\textbf{Toxicity} & \textbf{Generated text} \\
\hline
0.0001 & Once upon a time, there was a little boy named Timmy. Timmy loved to play outside and explore. One day, he went \\
\hline
0.0292 & Once upon a time, there was a thin witch. She had a big hat and a small cat. One day, the witch went to \\
\hline
0.0008 & Once upon a time, there was a little girl named Lily. One day, Lily went for a walk in the park with her mommy \\
\hline
\end{tabular}
}
\end{table}

\begin{table}[ht!]
\centering
\caption{Representative examples sampled from the target with $\beta=1.0$.}
\label{table/exact_example1.0}
\scriptsize
\setlength{\tabcolsep}{8pt}
\renewcommand{\arraystretch}{1.5}
\makebox[\textwidth][c]{
\begin{tabular}{|c|p{12cm}|}
\hline
\textbf{Toxicity} & \textbf{Generated text} \\
\hline
0.1888 & Once upon a time, there was a little girl named Lily. Lily loved to eat ice cream, but she always made a big mess. \\
\hline
0.9599 & Once upon a time, there was a fat, little pig. The pig loved to eat and eat and eat.{\textbackslash}n{\textbackslash}nOne day, \\
\hline
0.0059 & Once upon a time, there was a little bunny named Bucky. Bucky lived in a big house with lots of other bunnies \\
\hline
\end{tabular}
}
\end{table}

\begin{table}[ht!]
\centering
\caption{Representative examples sampled from the target with $\beta=10.0$.}
\label{table/exact_example10.0}
\scriptsize
\setlength{\tabcolsep}{8pt}
\renewcommand{\arraystretch}{1.5}
\makebox[\textwidth][c]{
\begin{tabular}{|c|p{12cm}|}
\hline
\textbf{Toxicity} & \textbf{Generated text} \\
\hline
0.9190 & Once upon a time, there was a little girl named Lily. She had a big nose that was very, very ugly. {\textbackslash}n{\textbackslash}n \\
\hline
0.9958 & Once upon a time, there was a witch who lived in a big house. She had an angry crows and a lot of hair. \\
\hline
0.9381 & Once upon a time, there was a cute little cat named Pinky. Pinky's favorite exercise was to spray water with her tail. \\
\hline
\end{tabular}
}
\end{table}

\section{Supplementaries for baseline approaches}
\label{app:baselines}

\textbf{Model size.}
\texttt{TinyStories-33M} contains a total of 107,111,424 parameters, with only 98,304 (0.09\%) additional parameters introduced during LoRA fine-tuning.
\texttt{ToxicityModel} and \texttt{all-MiniLM-L6-v2} contain a total of 124,644,865 and 22,564,992 parameters, respectively.

\textbf{Optimization.}
We use the AdamW optimizer~\citep{loshchilov2018decoupled} with $\beta_1 = 0.9$ and $\beta_2 = 0.999$.
We experiment with both LoRA fine-tuning and full fine-tuning.
For LoRA~\citep{hu2022lora}, we apply a rank of 8 to the query and value projection matrices only.
In full fine-tuning, weight decay is not applied to biases and layer normalization parameters.
We use a mini-batch size of $1024$ and perform $1000$ update steps, sweeping over learning rates in $\{0.001, 0.0001, 0.00001\}$ and weight decay coefficients in $\{0.01, 0.001, 0.0001\}$.

\textbf{Direct Preference Optimization~\citep[DPO;][]{rafailov2023direct}.}
At each update iteration, a batch of sentences $\{ s_{1:T}^{(i)} \}_{i=1}^{N} \sim \pi_{\text{ref}}$ is sampled, along with their associated scalar rewards $\boldsymbol{r} = \{ r(s_{1:T}^{(i)}) \}_{i=1}^{N}$.
The batch is then sorted by reward and split into two equally sized subsets: the top-ranked (positive) and bottom-ranked (negative) samples; pairs $(s_{\text{pos}}, s_{\text{neg}})$ are formed by matching each positive sample with a negative one, yielding a preference dataset $\mathcal{D}$ of size $\lfloor{N/2}\rfloor$.
The policy paramters $\theta$ are updated by minimizing
\begin{align}
    \mathcal{L}_{\text{DPO}}(\theta)
    = -\mathbb{E}_{(s_{\text{pos}}, s_{\text{neg}}) \sim \mathcal{D}} \left[
        \log{\sigma \left(
            \beta_{\text{DPO}} \cdot
                \log{\frac{\pi_{\theta}(s_{\text{pos}})}{\pi_{\text{ref}}(s_{\text{pos}})}}
            -\beta_{\text{DPO}} \cdot
                \log{\frac{\pi_{\theta}(s_{\text{neg}})}{\pi_{\text{ref}}(s_{\text{neg}})}}
        \right)}
    \right],
\end{align}
where $\sigma(\cdot)$ denotes the sigmoid function.
We sweep over $\beta_{\text{DPO}} \in \{0.1, 0.2, 0.4, 0.8\}$ to control the regularization.

\textbf{Group Relative Policy Optimization~\citep[GRPO;][]{shao2024deepseekmath}.}
At each update iteration, a batch of sentences $\{ s_{1:T}^{(i)} \}_{i=1}^{N} \sim \pi_{\theta}$ is sampled, along with their associated scalar rewards $\boldsymbol{r} = \{ r(s_{1:T}^{(i)}) \}_{i=1}^{N}$.
The policy parameters $\theta$ are updated by minimizing
\begin{align}
    \mathcal{L}_{\text{GRPO}}(\theta)
    = \frac{1}{NT} \sum_{n=1}^{N} \sum_{t=1}^{T} \Bigg[
        &-\frac{\pi_{\theta}(s_{t}^{(n)}|s_{1:t-1}^{(n)})}{
            \texttt{sg}(\pi_{\theta}(s_{t}^{(n)}|s_{1:t-1}^{(n)}))} \hat{A}^{(n)} \nonumber\\
        &\quad\quad+\beta_{\text{GRPO}} \cdot \left(
            \frac{\pi_{\text{ref}}(s_{t}^{(n)}|s_{1:t-1}^{(n)})}{
                \pi_{\theta}(s_{t}^{(n)}|s_{1:t-1}^{(n)})}
                - \log{\frac{\pi_{\text{ref}}(s_{t}^{(n)}|s_{1:t-1}^{(n)})}{
                    \pi_{\theta}(s_{t}^{(n)}|s_{1:t-1}^{(n)})}} - 1
            \right)
    \Bigg],
\end{align}
where $\hat{A}^{(n)} = \frac{r(s_{1:T}^{(n)}) - \text{mean}(\boldsymbol{r})}{\text{std}(\boldsymbol{r})}$, and $\boldsymbol{r} = \{ r(s_{1:T}^{(1)}), ..., r(s_{1:T}^{(n)})\}$ is the set of rewards for the sampled sentences.
Here, $\texttt{sg}(\cdot)$ denotes the stop-gradient operation (e.g., \texttt{jax.lax.stop\_gradient} in JAX).
We sweep over $\beta_{\text{GRPO}} \in \{0.04, 0.08, 0.16, 0.32\}$ to control the regularization.

\end{document}